\documentclass[5p]{elsarticle}

\usepackage{graphicx}
\usepackage[cmex10]{amsmath}
\usepackage{amssymb}
\usepackage{multirow}
\usepackage{color}
\usepackage{array}
\usepackage{times}
\usepackage{epsfig}
\usepackage{subfigure}
\usepackage{booktabs}
\usepackage{caption}
\usepackage{url}
\usepackage{hyperref}

\begin{document}

\begin{frontmatter}

\title{Traffic Lights with Auction-Based Controllers:\\Algorithms and Real-World Data}  

\author{Shumeet Baluja}

\author{Michele Covell}

\author{Rahul Sukthankar}
\address{Google, Inc.\\
1600 Amphitheatre Parkway\\
Mountain View, CA 94043\\}

\begin{abstract}
Real-time optimization of traffic flow addresses important practical
problems: reducing a driver's wasted time, improving city-wide
efficiency, reducing gas emissions and improving air quality.  Much of
the current research in traffic-light optimization relies on extending
the capabilities of traffic lights to either communicate with each
other or communicate with vehicles.  However, before such capabilities
become ubiquitous, opportunities exist to improve traffic lights by
being more responsive to current traffic situations within the
current, already deployed, infrastructure. In this paper, we introduce
a traffic light controller that employs bidding within micro-auctions
to efficiently incorporate traffic sensor information; no other
outside sources of information are assumed.  We train and test traffic
light controllers on large-scale data collected from opted-in Android
cell-phone users over a period of several months in Mountain View,
California and the River North neighborhood of Chicago, Illinois.  The
learned auction-based controllers surpass (in both the relevant
metrics of road-capacity and mean travel time) the currently deployed
lights, optimized static-program lights, and longer-term planning
approaches, in both cities, measured using real user driving data.

\end{abstract}


\end{frontmatter}


\section{Introduction}

Traffic congestion is a practical problem resulting
in substantial delays, extra fuel costs, and unnecessary harmful gas
emissions. In urban areas, traffic is largely controlled by traffic
lights.  Improving their control and responsiveness to existing travel
flows holds immense potential for alleviating congestion and its
associated problems.

\begin{figure}
\includegraphics[width=0.48\textwidth]{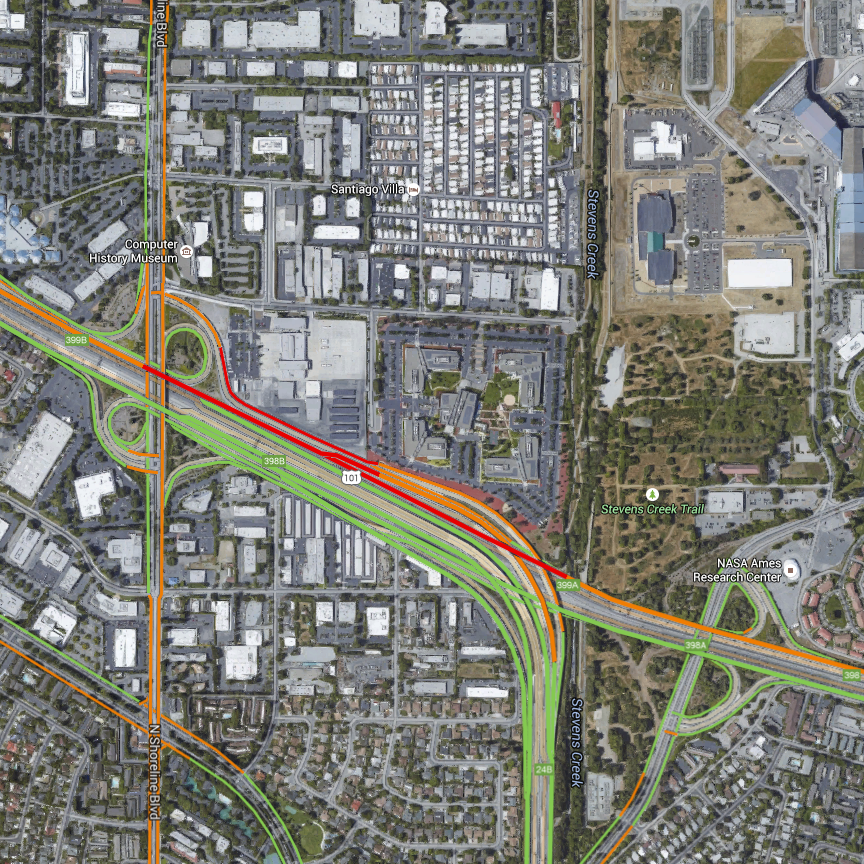}
\\
\\
\includegraphics[width=0.48\textwidth]{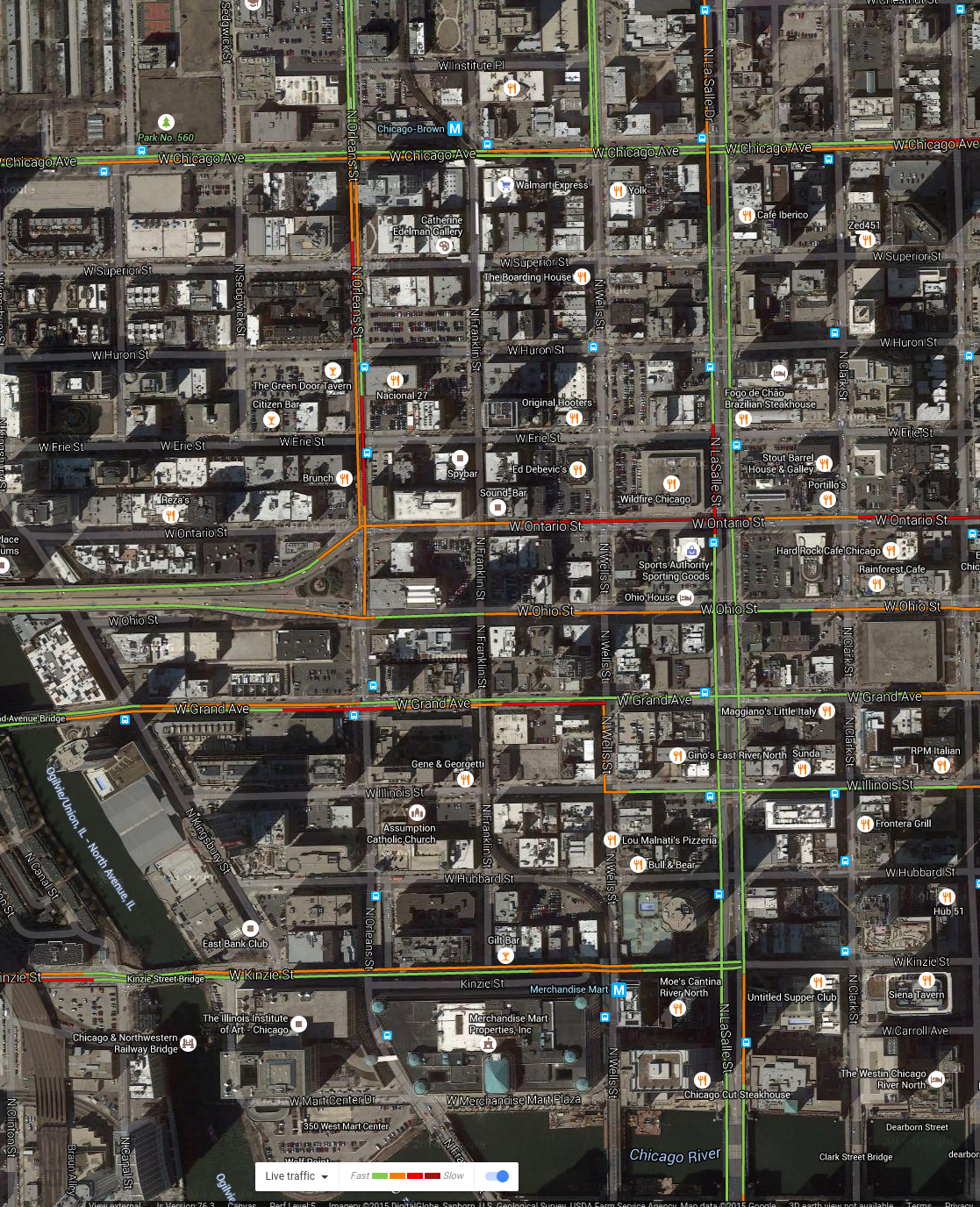}
\caption{ Top: Area to be optimized in Mountain View, California. A
  rush hour traffic flow shown (using Google Maps with Traffic
  Overlay.) Bottom: Area to be optimized in Chicago, Illinois. }
\label{figmtv}
\end{figure}

Inefficient configuration of traffic lights remains a common problem
in many urban areas -- one of the largest complaints of commuters in
the Mountain View, California area is the amount of traffic they face
during the morning and evening rush hours. One of the problem areas,
controlled by seven main lights, is shown in Figure~\ref{figmtv}
(Top).  The goal of our project is to evaluate the timing of the
traffic lights on these intersections and also improve them through
either better control algorithms or improved sensors.  For example,
many traffic lights are based on “fixed cycles”, which means that they
are set to green, yellow and red for fixed amounts of time.  Rarely is
this an optimal solution, as real-time traffic situations are not
considered, and can leave cars waiting in long queues to satisfy
shorter queues or even empty queues.  Nonetheless, even assuming
fixed-length, non-traffic-responsive lights, it is possible to
optimize the light timings to utilize historic knowledge of average
(or worst case) flows that have been observed.  Such approaches are
often tackled through the use of genetic algorithms to optimize the
light timings (phases and offsets)
~\cite{sanchez2004genetic,park2000enhanced,turky2009use,kalganova1999multiple}.
An alternate approach to the learning/optimization approaches of stochastic
search algorithms are methods that use \emph{reinforcement learning}
to discover optimal policies for the lights and/or cars
~\cite{wiering2000multi,arel2010reinforcement,moriarty1998,abdulhai2003}.

The methods referenced above have not only been applied to
fixed-policy lights, but also to lights that use vehicle sensors to
provide real-time traffic state information to the light controllers.
Additionally, many researchers have looked into future possibilities
where car-to-car, light-to-car, and car-to-light communication
exists~\cite{gradinescu2007adaptive, tielert2010impact,
  asadi2011predictive}.  Such communication allows for better light
control through the possibility of explicitly and directly
communicating light schedules to cars and car arrival times to lights.
Further, if we consider communication from lights to other lights,
such that each light could communicate its schedule as well as the
flows it is observing, more detailed planning and scheduling schemes
can be created~\cite{Smith2013,Xie2012-ICAPS,Xie2012-TRC}.  A good
overview can be found in~\cite{dresner2006}.

It should be mentioned that approaches relying on a centralized
controller or hierarchies of controllers have also been explored in
the research literature.  However, the more coordination that is
assumed, the greater the difficulties encountered in scalability.
Further, the problem of system ``nervousness'' grows when central
coordination is assumed --- small changes in the overall state of the
system may require large changes at the lower
levels~\cite{heisig2002nervousness}.  For scalability, we concentrate
only on local-decision making in this paper.

In the bidding-based traffic light controllers presented here,
external information is provided solely through \emph{local} sensors
(the most common being physically nearby in-roadway induction loop
sensors\footnote{Induction loop sensors are placed inside the
  roadway's pavement and, at a high-level, work by creating an
  electromagnetic field around the loop area. As vehicles enter and
  exit the field, fluctuations in the field are recorded as an
  indication that a car has passed over~\cite{WARD1985}.  They are
  widely deployed, and have been in use since the 1960s.}  or
cameras).  Rather than creating ad hoc rules for controlling
light-changes, the sensor information is placed within the framework
of a micro-auction.  When a light (phase) change is permitted, the
light controller collects \emph{bids} from all the phases and conducts
a micro-auction to determine the next phase.  Each phase's bids are
set by current readings from local induction-loop sensors.  The
bidding process and the weights of the bids are learned by our system
through the use of large-scale, real, historic data collected from
opted-in Android cell phone users.

This paper builds upon two branches of work: our initial studies with
micro-auction based controllers~\cite{covellTraffic2015} and the
data-driven estimation of deployed traffic lights~\cite{Baluja2015}.
To avoid potential confusion, it should be remarked that the auctions
in this paper substantially differs from auction-related approaches in
which drivers and/or automated cars bid for the right of
way~\cite{carlino2013,schepperle2007,schepperle2008traffic} to gain
favorable light timings.  In this work, the micro-auctions serve as a
unifying \emph{internal} mechanism to handle the complexities of
prioritizing the different phases (colors/settings) of the lights.

There are two primary contributions of this work. The first, as
described above, is the introduction of an auction-based controller
that relies primarily on currently deployed sensors and does not
assume the existence of future communication between sensors. The
proposed controller is tested on months of real data in two large
cities.  The second contribution of this paper is to provide a novel
method for establishing a realistic baseline of performance based on
the \emph{currently deployed} traffic controllers.  This addresses an
often overlooked pragmatic issue --- in many real environments, access
to the programs of the existing light deployed throughout cities is
not available.  Without this information, however, realistically
assessing improvements over baselines is impossible.  We address this
problem through a novel method of creating models of the behavior of
the currently deployed lights based on observed traffic data.  This
yields a realistic baseline of performance with which to compare
proposed improvements --- including those of our own traffic
controllers.

To provide the necessary background for understanding traffic light
control, in the next section, we briefly describe how to simulate
unconstrained transitioning between light phases, which is fundamental
for fully responsive controllers.  We also describe a powerful recent
state of the art planning-based traffic controller ~\cite{Smith2013}
which was has been successfully deployed in Pittsburgh, Pennsylvania.
This provides a strong competitor to our approach.
Section~\ref{auction} details our micro-auction controller.  All of
the methods presented in this paper require learning many
parameters. A simple learning/optimization procedure, termed
\emph{Next Ascent Stochastic Hillclimbing (NASH)}, is presented in
Section~\ref{data}.  Experiments with alternative optimization
procedures, such as Genetic Algorithms and Probabilistic Model Based
Optimization, are also described.

As mentioned earlier, a \emph{real} baseline of performance, one which
considers the programs/schedules of the currently deployed lights, is
difficult to obtain.  Before we present our final results, in
Section~\ref{estimation}, we describe a practical method to estimate
the behavior of the currently deployed traffic light schedules based
on collected traffic data.  Although this section is not part of the
auction-based algorithm itself, the procedures presented in this
section are vital to accurately estimate the \emph{real} expected
differences in performance over current traffic-light programs.

With the baselines obtained through the methods described in
Section~\ref{estimation}, we can provide confident measurements of the
effects of bidding-based traffic controllers.  Large scale,
extensive, empirical results are presented in Sections~\ref{mtv}
\&~\ref{chicago}, for Mountain View California and Chicago, Illinois.
The auction-based controllers yield improvements in the both the
important measurements of road-capacity and mean travel time within
both urban environments.  The paper concludes with directions for
future work in Section~\ref{concs}.

\section{Background Context: Traffic Light Control with Unconstrained Transitions}

\begin{figure}
\centering
\includegraphics[width=0.3\textwidth]{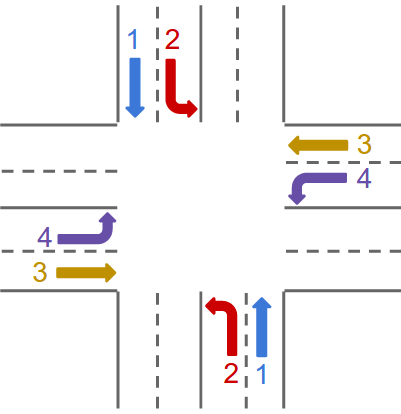}
\caption{An example four-phase traffic light}
\label{intersection}
\end{figure}

At a high level, whenever there is a transition from one traffic-light
phase to another, where some of the lights that were green become red,
there is the need for an intermediate phase, a \emph{yellow phase}, to
provide warning to the affected drivers to stop or to proceed, based
on their speed and distance to the intersection.  In static traffic
lights, the sequence of green phases is simply round-robin (e.g., in
Figure~\ref{intersection}, repeatedly going through the four indicated
green phases in counting order).  In the case of round-robin
transitions, since the sequence of greens is fixed and pre-defined, it
follows that there is only one yellow phase that will follow any given
green phase.  However, when we employ traffic sensor inputs, we are
given the flexibility for out-of-sequence transitions (e.g., phase~1,
then phase~3, and so on), based on the readings of the current
intersection's local induction loops.  This may yield improved traffic
flow, but it also introduces the need for additional yellow-phase
logic.

In our experiments, both with planning-based lights and micro-auction
based lights, in order to be responsive to sensed-traffic, we allow
lights to make out-of-sequence transitions between green phases.  This
allows more complete use of the intersection capacity.  In addition,
when the traffic-light logic includes advance planning, there is a
second advantage to out-of-sequence transitions by shortening the
planning horizon. With planning-based logic, forcing a round-robin
sequencing necessitates long planning horizons to avoid making
short-sighted decisions --- the light must plan ahead for the combined
duration of all the phases in the cycle.  In contrast, with
unconstrained sequences, the light only needs to plan ahead for the
duration of a single yellow phase.

Which of the signal lights needs to be yellow depends on the specific
combination of the previous and subsequent green phases. Since we
allow out-of-sequence transitions between green phases, we create a
grid of the correct yellow phases to place between each transition.
During the simulation setup~\cite{Krajzewicz2012}, for each pair of
green phases, if any green state needs to change to a red state, the
intermediate yellow phase is simply constructed by setting those
soon-to-be-red states as yellow.  The second step is setting the
duration of the yellow.  We can look at the states that are marked as
yellow and trace backwards from them to the maximum of the speed
limits for the lanes that those signal states control. The yellow for
that pair is set to the maximum speed divided by the Department Of
Transportation (DOT)-recommended safe deceleration rate of 3
m/s$^{2}$~\cite{DOT2013} plus a reaction time of 1 second.

As an implementation note, we remark here that all experiments were
conducted within the SUMO~\cite{Krajzewicz2012} simulator.  SUMO
(\emph{S}imulation of \emph{U}rban \emph{MO}bility) is a traffic micro
simulation package~\cite{SUMO2015}, that uses discrete time steps (1
msec each) in its simulations but keeps a continuous representation of
location, distance, and speed.~\footnote {Two important considerations
  in selecting SUMO were: (1) SUMO is written in a combination of C++
  and Python that can be easily modified; this was important for being
  able to handle non-standard light logic.  (2) The free license
  allows us to parallelize it on multiple machines; in our tests,
  thousands of tests have been carried out simultaneously --- allowing
  deeper exploration than would be possible with more restrictive
  licenses.}  This continuous spatial representation improves
simulations that include congested surface streets, compared to many
discrete-space alternatives~\cite{Tonguz2009}.

\subsection{Prior Work: Long-Term Planning Based Approaches}

The micro-auction based traffic lights that are the focus of this
paper are \emph {reactive} by design; they do not have a
\emph{planning} mechanism for future events.  Complementary
approaches, such as planning based traffic lights that attempt to
anticipate the traffic that will arrive in the future, provide
additional avenues to explore.  We have implemented planning based
lights to compare their effectiveness with micro-auction lights.  One
implementation of planning based lights is described here.

Inspired by the exemplary results of the Carnegie Mellon University
Pittsburgh traffic-light deployment~\cite{Smith2013}, we implemented a
traffic light controller that uses remote sensors and planning for
controlling when traffic phase switches occur.  With the
planning-based approach, we solve the phase scheduling problem by
collecting the traffic data from a sequence of induction loops that
are tied to a given phase of the traffic light, with the loops spaced
at about 3 seconds of expected travel time apart from each other.  The
sensors are placed in the roadway, spaced at distances dictated by the
speed limit and this target separation time.  In the planning-based
approach, we use these sensors for both occupancy and speed
data~\cite{Thakuriah2013}.  We assume that they have been placed on
all lanes that could lead traffic to the controlled intersection in 15
seconds (or less) of travel.  When intersections are closely spaced,
these non-local sensors may be on the other side of other
intersections, leading to a large web of sensors and distributed
communication.  While this likely includes more sensors (and
communication) than is minimally required, it enables us to create a
detailed speed and occupancy profile for each section of road, which
is useful for planning.

Internal to our planning-based traffic lights, we maintain time lines
of when we expect cars (observed by local and remote induction loops)
to arrive at the controlled intersection.  The car counts (from
induction loops) are scaled by a learned weighting factor, based on the
historically observed turning ratios between the sensor and the
traffic light, as well as the historically observed rates for which
the phase will be needed by the incoming traffic (e.g., straight or
left turn at the controlled intersection itself).  The
distance-to-time mapping used to put the car onto the time line is
based on the observed speed profile for the road (also from the known
induction loops).  Finally, we use dynamic programming to solve for
the best phase sequence and transition times.  Since this is a crucial
part of the planning process, it is described in more detail.

The dynamic-programming search is initialized with a single potential
schedule that has the start time of the current phase as its (already
past) starting point.  If there are cars waiting at the traffic light
on other phases, it increases the solution space by considering a
phase change to each of the other phases that are in demand, taking
into account the yellow-duration lead time needed for scheduling.  For
example, considering Figure~\ref{intersection}, if phase 3 is the
current phase and there is traffic waiting for phase 1 and phase 2,
the space of possible solutions will expand to three possibilities:
(a) remaining on phase 3; (b) changing to phase 1; and (c) changing to
phase 2.  This expansion of the possible solution space continues with
new possible branches being introduced each time a new car arrives.

The timing of the phase changes is restricted so that the minimum time
given for each phase of the light (plus the yellow duration) is
respected: e.g. if phase~4 requires a minimum duration of 3~seconds,
then the next possible phase change will be postponed until 3~seconds
plus the yellow duration after the start of phase~4.  Also, if the
current phase is expected to be ``empty'' before the arrival of the
new traffic (that is, no cars waiting as measured by the local
induction loop sensors and no cars arriving as measured by the remote
induction loop sensors), the proposed phase change for the newly
arriving traffic is moved up to be as early as possible, such that the
yellow will start just after the current phase is empty.  This early
change has the advantage of reducing the amount of time that the
intersection remains under utilized.

Given the complete description of how phases can change, the
scheduling solution is then selected based on minimizing a combination
of three penalties (the automatic setting of the weights of these
penalties is described in Section~\ref{data}).

\begin{enumerate}
\item Speed-loss penalty: This is the penalty for forcing a car to
  stop.  The penalty is scaled by the speed limit of the road that the
  car will be traveling onto.  That speed was chosen since it best
  reflects the acceleration that the car will require, that it
  otherwise would not have needed had it not been forced to stop.
\item Waiting-time penalty: This is a penalty that is linear in the
  amount of time that each car must wait at a red light.
\item Phase-change penalty: This is a penalty that is added for each
  proposed phase change.  It increases the penalty on the schedules
  that unnecessarily cycle through the different phases when there is
  no waiting or incoming traffic.
\end{enumerate}

The solution space is repeatedly expanded and then pruned, using the
approach described in~\cite{Xie2012-TRC}.  It is expanded by moving
forward through all of the arrival times of incoming traffic.  After
each expansion, the solution space is checked for schedules that can
be pruned.  Pruning of candidate solutions happens with solutions that
end on the same phase and have the same number of waiting cars as
other solutions, but have higher associated partial costs (longer
times).  Experiments with the planning-based controllers are presented
in Section~\ref{mtv}.

In contrast to the approaches described in this section, the
alternatives that we propose, auction-based controllers, do not create
plans and are purely reactive.  They are described next.

\section{Auction-Based Controllers}
\label{auction}

\begin{figure*}
\centering
\includegraphics[width=1.0\textwidth]{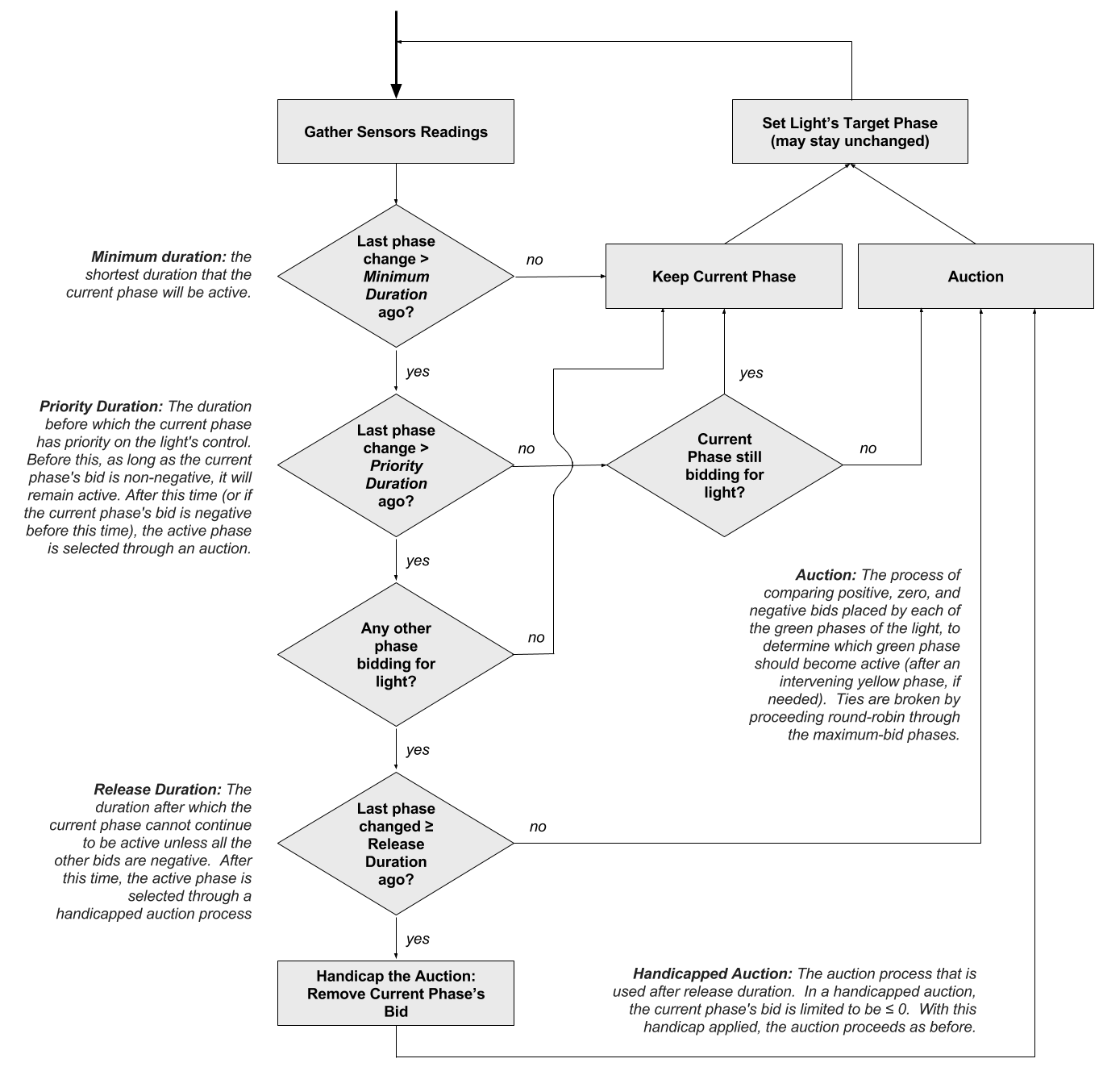}
\caption{Phase change logic in auction-based traffic lights.}
\label{flowgraph}
\end{figure*}

In this section, a detailed description of the new auction-based
controllers is provided.  In contrast with the planning-based
traffic-light control~\cite{Xie2012-TRC}, the auction-based approach
does not require the use of \emph{remote} sensors since no planning is
required.  Instead, only the typical induction loops that are placed
at the entrance lanes to the controlled intersection are used.
Because we rely only on existing roadway infrastructure, these
controllers should be easier to deploy than those that require remote
communication.

At a high level, each phase of the auction-based traffic light logic
has three time-separated behaviors (see Figure~\ref{flowgraph}). The
timing and inputs used for these behaviors are optimized to the
general traffic patterns that are expected a given time of day (e.g.,
morning commute hours).  Based on our tests, it is the combination of
these three behaviors along with the parameter optimization that are
responsible for the improved efficiency witnessed over static timed
and planning-based approaches.  The behaviors are described below and
the parameter optimization is explained in the next section.

In auction-based logic, each traffic-light phase definition includes a
weighted list of sensors that is to be used by that phase to determine
its \emph{bid} for the cycle at any given time.  The weights can be
positive or negative and are used to effectively scale the number of
cars observed on that sensor.  For example, a phase for a
lower-priority road could include in its sensor list a
\emph{negative-weighted} induction loop from in front of the
higher-priority road, so that the lower-priority road would be more
likely to release the phase when cars arrive from the higher-priority
direction.  Similarly, a more heavily-used phase could use a larger
positive weight for some of its own sensors, to allow it to
\emph{out-bid} the lighter-traffic direction.  Instead of having these
weighted lists of sensors be given as a fixed input that is manually
specified, the learning procedure selects both which of the local
sensors to use and their associated weight \emph{for each phase}.
Formally, a bid is the weighted sum of the current, local
induction-loop measurements ($s_j$): the bid $b_i$ for phase $i$ is
$b_i = \sum_{j} w_{ij} s_j$.  The weights ($w_{ij}$) are selected in
the learning process (see Section~\ref{data}).  When the learning
process sets $w_{ij} = 0$, we refer to that as having removed sensor
$j$ from phase $i$.

As shown in Figure~\ref{flowgraph}, the way in which the traffic light
decides whether to change phase depends on how long the current phase
has been active.  We use the terms \emph {minimum duration},
\emph{priority duration}, and \emph{release duration} to separate the
time intervals for these different behaviors.  As suggested by its
name, minimum duration is the minimum amount of time that a phase must
be green before possibly changing to yellow.  The minimum duration is
given as an input to the simulation and can be set to be different for
each phase of the light.  For simplicity, we start with all of the
minimum durations as 3 seconds but allow the learning procedure to
adjust that to be larger, if needed to avoid too-fast switching
between phases in light traffic.  No phase changes can occur before
minimum duration.

For each second between minimum duration and \emph{priority duration},
the current phase has priority on the traffic light.  If its bid for
the cycle is non-negative (indicating that it would like to have the
cycle), then it will keep the cycle, no matter what the bids of the
other phases are.  While this greedy approach may seem to be
suboptimal, it has the advantage of increasing the average duration of
the cycles and reducing the amount of time spent switching between
green phases (and thereby reducing amount of time wasted on yellow
lights).  Again, we allow the learning procedure to adjust the
priority duration to the expected traffic demands.

For each second between priority duration and \emph{release duration},
an auction is held between the different phases.  Each phase bids
according to the weighted sum of the sensors that have been selected
for that phase.  If the highest bid is negative, then the current
phase is the default winner of the cycle until one or more of the bids
change.  Otherwise, the phase with the highest bid will get the cycle
(after the appropriate yellow).  If multiple phases have the same
winning bid, the winner is selected by simple round-robin selection.

For each second after the release duration, it is important to make
phase switching less restrictive. The same type of auction is held
with the added constraint that the current phase cannot bid an amount
above zero. This non-positive bid by the current phase likely releases
the cycle.  Any other phase that would like to have the cycle will win
the auction away from the current phase.  If more than one of the
other phases have positive bids, the auction process will pick the
strongest bidder.  Only if all the bids are strongly negative, is it
possible for the current phase to stay green.

The progression restarts after the start of each green phase,
progressing from non-negotiable (below minimum duration) to greedy
(below priority duration) to auctioned (below release duration) to a
handicapped auction (above release duration).

Because the learning procedure is allowed to remove any or all of the
sensor inputs to any given phase’s bid, we designed our control logic
to handle these cases. When a phase has no sensor inputs, it will
continually place a zero bid for the light.  If the learning procedure
removes all sensor inputs from all phases of the traffic light, the
auction logic results in the light behaving as a static light, using
round-robin cycling and using each phase's \emph{priority
  duration}. Interestingly, in practice, the no-sensor configuration
is automatically instantiated through our learning procedure for
multiple lights; this will be discussed in the results sections.

\section {Learning/Optimization Algorithms and Data Gathered}
\label{data}

No matter which approach is used for traffic light control,
fixed-schedule, long term planning, or micro-auctions, each has
numerous internal parameters that must be set.  For example, even with
a simple fixed-schedule controller, the length of the phase and
offsets of each light have a large impact on the performance of the
overall system.  Table~\ref{params} summarizes the parameters that
characterize the behavior for a given traffic light in each of the
three approaches explored here.  Genetic Algorithms
(GA)~\cite{goldberg1989genetic} have been most commonly used to set
the numeric and enumerable values associated with traffic
lights~\cite{sanchez2004genetic,park2000enhanced,turky2009use,kalganova1999multiple}.

\begin{table*}
    \centering
    \caption{Parameter Categories Optimized for Each Approach}
    \label{optimized parameters}
    \begin{tabular}{p{4cm}p{6cm}p{6cm}}
    \toprule
    Fixed-Schedule& Planning & Micro-Auctions \\
    \midrule
    Phase Lengths & Speed Loss Penalty Weight  & Detector Weights\\ 
    Phase Offsets & Waiting Time Penalty Weight  & Detectors to Use \\ 
     & Phase-Change Penalty Weight & Durations (Minimum, Priority, Release) \\ 
    \bottomrule
    \end{tabular}
  \label{params}

\end{table*}

Following the published research, genetic algorithms were first
attempted.  A thorough overview of GAs can be found in
~\cite{goldberg1989genetic}. Since genetic algorithms are themselves
characterized by a number of control parameters, over 40 different
combinations were tried.  In the GA variants explored, the control
parameters that were varied included: the mutation rate was varied
between 0.5\% - 10\%, crossover type employed: uniform and
single/multi point, the use of elitist selection (preserving best
solution from one generation to the next), population sizes were set
between 10-1000, and numerous crossover rates were attempted.  Two
settings consistently had the largest impact.  The largest effect on
the performance of the algorithm was observed in the setting of the
mutation rate.  Regardless of the other parameters, the entire
population converged to far from-optimal candidate solutions that were
then substantially improved through random mutations.  Even when
crossover was turned off (probability = 0.0\%), the results were not
statistically different than when crossover was used.  The other
control parameter, whether elitist selection was turned on, was
important.  In the runs where it was used, the final solutions found
were consistently better than when it was not used.

In addition to GAs, probabilistic optimization approaches such as
Population Based Incremental Learning (PBIL) \cite{baluja1997}, and
variants that include explicit modeling of inter-parameter
dependencies were also attempted~\cite{baluja1998,harik1999}.  These
probabilistic models provide a method to explicitly maintain the
statistics that a genetic algorithm's population implicitly
maintains. A good overview of probabilistic optimization techniques is
provided in~\cite{pelikan2002}.  For our experiments, the
inter-parameter dependencies that were modeled were pair-wise
dependencies using tree-shaped networks.  As shown in~\cite{Chow1968},
a simple algorithm can be employed to select the optimal tree-shaped
network for a maximum-likelihood model of the data (in our case, the
data is the high performing parameter settings found through search).
This tree-model is then sampled to generate the next candidate
solution in an analogous step to the crossover operation in a GA.
Like the GAs attempted, a large driver of improvement was the mutation
operator. ~\footnote{Brief attempts with other heuristic optimization
  procedures were also conducted.  Although they sometimes
  outperformed the GAs used, they did not provide any statistical
  benefit over simple hillclimbing.  For simplicity, we use only
  hillclimbing throughout the remainder of this paper. }

Based on the consistently large effects of the mutation operator on
the quality of the solution obtained, a simpler search method
\emph{Next-Ascent Stochastic Hillclimbing (NASH)}, was tried.  It is
described below.

Formally, we define $S$ to be the set of parameters that fully specify
a given scenario.  For instance, a micro-auction simulation containing
10 traffic lights would require setting $|S| = 50$ parameters (5
parameters per light).  Since different lights will (in general)
employ different parameter settings, their behavior will not be
identical.  The goal of our optimization is to identify parameter set
$\hat{S}$ that minimizes the total travel time for the cars in the
simulation (detailed more formally below).  \emph{NASH} operates in a
manner similar to other stochastic hillclimbing methods:

\begin{enumerate}

\item \textbf{Perturb $S$:} Since $S$ is an aggregation of parameters, we randomly select a parameter $s \in S$ (drawn with uniform probability) to perturb as:
\[
\begin{array}{ll}
s' \leftarrow s + U(-0.05 s, 0.05 s)	& \text{for continuous-valued $s$;} \\
s' \leftarrow U(\text{options}(s) \setminus s)	& \text{for discrete-valued $s$.}
\end{array}
\]
where $U(.,.)$ denotes a draw from the uniform distribution and $\text{options}(s)$ denotes the set of legal values for a discrete parameter.  In other words, we perturb continuous parameters by $\pm 5\%$ and assign a random legal value to enumerable parameters.
We do not simultaneously perturb every parameter in $s$; the number of parameters changed in one operation is given by sampling $U(1, 0.05 |S|)$ (i.e., perturb no more than 5\% of the total parameters in the simulation in each iteration). The perturbed parameter set is denoted as $S'$.

\item \textbf{Project $S'$ onto valid subspace:} Since the
  perturbation operation described above operates independently over
  parameters in $S$, it can generate a setting $S'$ that violates
  constraints over parameter subsets. For instance, increasing a phase
  length in a given fixed-schedule traffic light controller will break
  synchronization unless another phase length for the same light is
  reduced to keep the overall cycle constant. To address this, we
  employ an approach-specific \emph{repair} operation over $S'$ to
  adjust parameter settings (e.g., normalizing parameters for each
  light).

\item \textbf{Evaluate objective:} With the parameter modifications,
  the new $S'$ is evaluated with the desired \emph{objective
    function}.  The standard objective functions that have been used
  in the literature include: minimizing the overall or average wait
  time, maximum wait time, emissions, stop time, or maximizing
  throughput, speed, etc.  For our studies, we set the objective
  function to minimize total travel time of all the cars that enter
  the region of interest  during the
  observation interval.

  Our implementation assigns a pre-generated route and entry time for
  each car $c \in C$ in the simulation. These are drawn directly from
  the (anonymized) real gathered data of Android users, as will be
  described next, in Section~\ref{anonymousdata}.  The exit time for
  the car depends on the traffic flow in the region of interest; in
  general, poor settings of $S$ result in the car exiting the region
  of interest later.  Thus, the objective of total travel time is
  given by
  \[
  	O = \sum_{c \in C} c_\text{exit} - c_\text{entry}
  \]
where $c_\text{entry}$ and $c_\text{exit}$ denote the entry and exit times for car $c$.  Since $\sum_{c \in C} c_\text{entry}$ is constant across iterations (deterministic arrival times and routes), and $c_\text{exit}$ is determined by the parameter settings $S$, our optimization problem can be concisely stated as
  \[
	\underset{S}{\text{minimize}} \sum_{c \in C} c_\text{exit} 
  \]

\item \textbf{Accept S':} If the perturbed parameters $S'$ result in an improvement over $S$ on the objective, we accept the perturbed settings,
$S \leftarrow S'$. Otherwise, we discard $S'$.

\end{enumerate}

This process is iterated until either a satisfactory solution is found
or time expires.  Though extremely simple, NASH worked as effectively
as the other optimization methods tried, consistently outperforming
GAs. Further, it was simpler to implement and faster in practice than
GAs and building probabilistic models.  This somewhat
counter-intuitive result has also been observed by other researchers
in exploring the trade-offs between genetic algorithms and stochastic
hillclimbing techniques~\cite{Juels1995,mitchell1993,harman2007,rosete1999}.
This finding is
especially pronounced in problems in which mutation (as opposed to the
Genetic Algorithm's primary search operator, crossover) is the main
driver for improvement in the solution --- as we have empirically
found to be the case in this problem.

We employed a straightforward objective function that meets the goal
of minimizing overall travel times.  However, it is worth noting that
many other objective functions could have been used.  Two common ones
include minimizing driver idle or wait-times and minimizing emissions.
Although both of these can be used here with no change to the
algorithms or optimization procedures (other than the objective
function); their effects can be large.  For example, in minimizing
emissions, reducing stop times at lights and consistency of speed
may take higher precedence than minimizing overall travel time.
Although this may also reduce overall travel time, that reduction will
be a side-effect of reducing the emissions.  Additionally, other
interesting, more specialized objective functions can be used as well,
such as minimizing travel times on specified roads or routes, or even
minimizing congestion at a specific light.

\begin{figure*}
\centering
\includegraphics[width=1.0\textwidth]{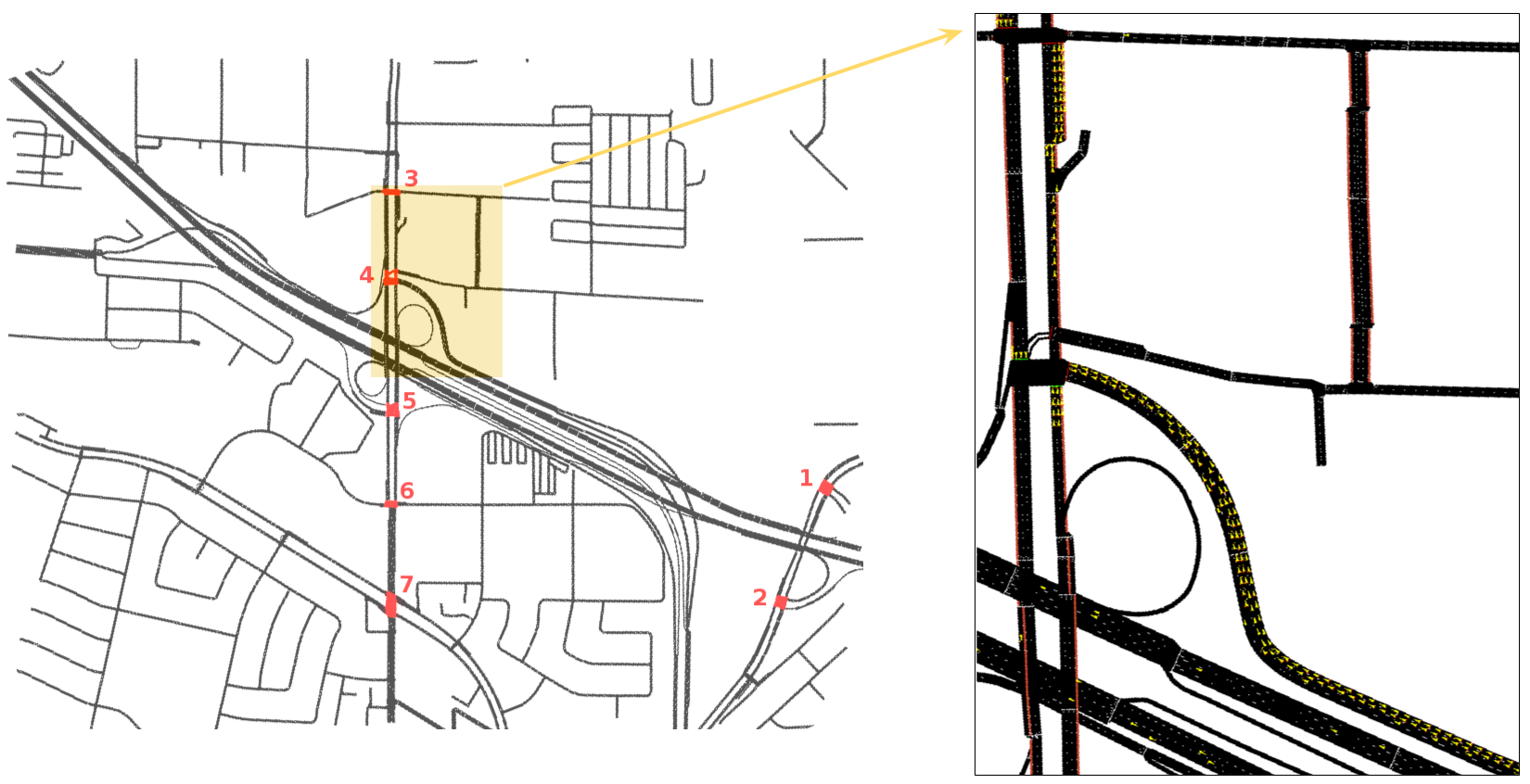}
\caption{Roadway data imported into SUMO
  simulator~\cite{Krajzewicz2012}.  Left: Overview of the area we are
  considering in Mountain View, CA. The seven lights considered are
  numbered (and appear in red).  Right: Enlarged section with traffic
  at Shoreline Blvd. and 101/85 exit ramps at 9am on a summer Tuesday.
  Each yellow triangle represents one vehicle.}
\label{fig:mtv-sumo}
\end{figure*}

\subsection{Anonymous Real-World Traffic Data}
\label{anonymousdata}

For the real-world experiments presented in this paper, two sets of
data are needed: the roadway information (layout, speed limits, etc.),
and travel-track information.  To gather the road information, we
combined the data available from Google maps and
OpenStreetMap~\cite{Haklay2008}.  The results provided roadways as
well as traffic light locations, as shown in
Figure~\ref{fig:mtv-sumo}.

In addition to accurate road information, demand for each road section
must be modeled.  We created a demand profile through anonymized
location data collected from opted-in Android cell phone users
~\cite{Barth2009}\cite{Dobie2014}. The data was collected over several
months.  The raw data, which itself does not include personally
identifiable information, was additionally scrubbed by segmenting the
travel-tracks to prevent association of trip origins and destinations
to even further reduce identifiability risks.  See
Figure~\ref{fig:mtv-sumo}(Enlarged Section) for a sample of the data
collected.

From this data, we further sub-select the data with travel-tracks that
intersect with the map area shown.  We also filter by time, limiting
to looking at a given start and end date --- in particular around rush
hour periods.  We then filter all the given times down to the weekday
and time of day (e.g., Tuesday 7am local time).  This provides a
close-to-realistic profile of the road demand for Android users.  By
\emph{folding/overlapping} the gathered data over weeks, we
compensated for the fact that all travelers are not Android users who
provide their data.

\section{Establishing a Baseline: Modeling the Behavior of Currently Deployed Lights}
\label{estimation}

Before conducting the experiments, we need a method of determining
whether the new controllers will improve real traffic flow beyond the
currently deployed traffic light controllers.  In this section, we step
away from discussing algorithms for traffic light control and examine
the pragmatic issue of how to obtain the existing traffic light
schedules and behaviors from which to create a realistic baseline of
performance.

Unfortunately, for deployed traffic lights, the running light
schedules/programs are often not known.  Rarely are they kept in a
central database, and even when they are, they are often not easily
obtainable.  The most straight-forward solution is to manually watch
and time each traffic light in the city to be optimized.  For lights
on fixed schedules this may be theoretically feasible, but will likely
be prohibitively expensive and time-consuming to do at scale.
Nonetheless, without the existing schedule information, it is
difficult to ascertain any real-improvements that new traffic light
algorithms and approaches will have in reality.

We take a \emph{behaviorist} approach to discover lights'
\textbf{currently deployed schedules}. We attempt to discover the
light schedules for all the lights in the system we wish to model.
This is done by creating models that match known car travel paths and
timings that were observed in a large collection of gathered
travel-tracks.  If we can determine the lights' programs accurately,
when simulating the known travel-tracks through the city, the timings
will match observed timings.  With this derived model, we can simulate
their performance 
light's programs with varying traffic conditions to test how they will
scale in comparison to any of the alternative light controllers that
we propose.

Unlike the rest of this paper, in which which we use \emph{NASH} to
optimize controllers to improve traffic lights' schedules, in this
section, the goal is to obtain a light schedule that matches, as
closely as possible, the behavior of the current pre-existing
schedule.  Recall that \emph{NASH} relies on a repeated stochastic
generation and evaluation methodology to guide search.  The evaluation
of the candidate traffic light schedules is controlled by setting an
\emph{objective function} (in the description of the \emph{NASH}
algorithm presented in Section~\ref{data}, see step 3).  Instead of
using the commonly employed traffic-optimization objective functions
to improve some aspect of flow throughput, we specify the following
objective function based on similarity (formulated as a minimization
task):

\begin{equation}
f(h)= \sum_{c \in \cal C}
  |\text{JourneyTime}_{h}(c) - \text{JourneyTime}_{a}(c) | ,
  \label{eqn:obj}
\end{equation}

where $h$ and $a$ are the hypothesized and actual light settings,
respectively, and $\cal C$ denotes the set of cars in the simulation.

Minimizing this objective allows us to determine light settings that
generate a simulated traffic flow that closely mirrors the actual
flow.  Note that although the constraint is not explicitly specified
in Eqn.~\ref{eqn:obj}, the cars are all introduced into the system in
the same order and at the same times as they were observed in the
actual data; this is crucial to ensure that traffic jams and roadway
usage are emulated correctly.  As a reminder, note that this
similarity measurement is suitable only for determining the currently
deployed light schedules.  Similarity is \emph{not} the correct
measurement to use when optimizing the light controllers to minimizing
wait time, emissions, total travel time, etc.  For those objectives,
we will simply minimize the specific objective instead.

\subsection{Validating the Estimates on Controlled Data}

\begin{figure}
\centering
\includegraphics[width=0.49\textwidth]{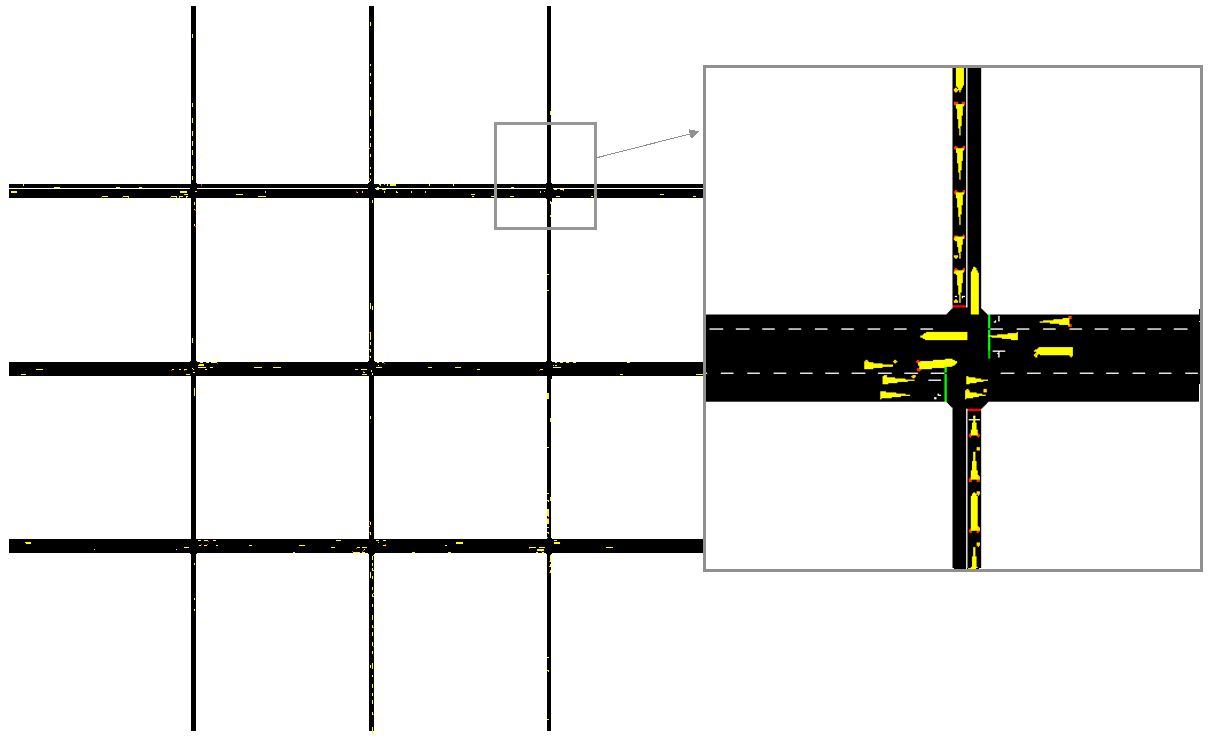}

\caption{ The synthetic roadway.  The number of lanes varies between 1
  and 3 in each direction.  Cars can be introduced and exit at any of
  the twelve outer edges.  Lights are at located at each of the nine
  intersections.  Enlarged region shows typical traffic. }

\label{fig:synth}
\end{figure}

Because estimating deployed traffic light parameters is a novel part
of this study, we validate the approach with a series of experiments
using synthetic data before turning to real-world data.  Using a
smaller, synthetic, data set has the advantages of being noise-free
and completely within our control to modify in order to examine
different aspects of matching.  For the synthetic data, 3200 cars were
instantiated over a period of 4000 seconds to travel along the simple
grid shown in Figure~\ref{fig:synth}.  The speed limit for each
segment was chosen independently and randomly, and seven types of car
were instantiated, with differing profiles in terms of
acceleration/deceleration, following distance, length, etc.

Each simulated car's path was chosen to start and end at randomly
selected edge nodes. The path was constrained such that no
intersection was visited twice.  Vehicle launch times were uniformly
randomly distributed over the 4000 seconds.

The 3200 cars were simulated in SUMO with a
pseudo-randomized~\footnote{This light setting was significantly
  perturbed from SUMO's default light setting to ensure that it could
  not ``accidentally'' be found by NASH by initializing SUMO and
  making tiny perturbations to their default light settings.} light
setting, this is referred to as the \emph{Target-Light-Setting}.  For
the synthetic data experiments, the \emph{Target-Light-Setting}
corresponds to the settings for traffic lights that we seek to
estimate using the procedures in this section.  For simplicity of
exposition, here we assume that the actual traffic lights can be
modeled with fixed schedules.  As discussed later, more complex
controllers that incorporate induction loop sensors also fit naturally
within this procedure.

\emph{NASH} was then applied to the uncalibrated, randomly
initialized, lights to adjust them such that the cars had
approximately the same travel times as in the original distribution.
The objective function was to match, for each travel path, the arrival
time of the car as closely as possible to what SUMO yielded with the
\emph{Target-Light-Setting}.  Recall that because we want to emulate
not having \textbf{any} \emph{a priori} information about the actual
light settings, we start NASH with random settings for all the lights
in the system.  The hope is that through learning, the light settings
that are found will behave the same as those in
\emph{Target-Light-Setting}.  It is interesting to note that there may
be many possible light settings that generate similar aggregate
traffic flow behaviors.  As will be shown in later experiments, when
multiple NASH runs are conducted, different, but equally well
matching, light settings are found.

\begin{figure}
\centering
\includegraphics[width=0.46\textwidth]{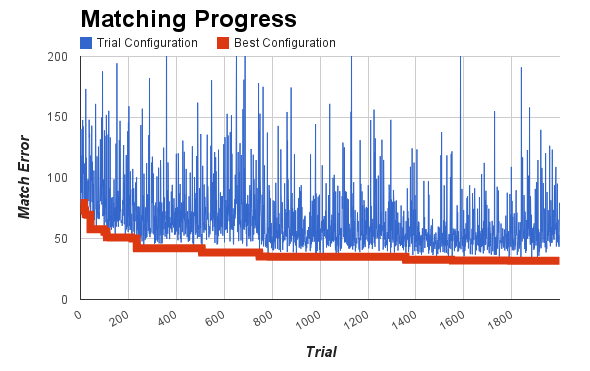}
\hfill
\caption{Match Error over successive trials. 2000 trials shown.
  Y-axis the average difference in journey times for cars under the
  hypothesized light setting and the actual light setting (in
  seconds).  X-axis: trial number.}
\label{fig:optSynth}
\end{figure}

The progress of the matching algorithm is shown in
Figure~\ref{fig:optSynth}.  Two lines are shown, the bottom line (in
red) shows the best light setting that was found in the search to that
point. The top line (in blue) shows the evaluation of each candidate
light setting as search progresses.  From the red line, observe that in
the beginning of learning, the average difference (in seconds)
between when a car reaches its destination with the original
\emph{Target-Light-Setting} and the approximated light-setting was
over 80 seconds.  By the end of learning, this is reduced to 31.8
seconds.  Also notice that the blue line is riddled with spikes.  This
means that the majority of perturbations to the best-light setting
found to that point yielded matches that performed significantly
worse; small mutations in the light settings caused drastic changes in
the overall traffic flow. Only a few trials, those where the red line
took a step downwards, revealed an improved performance.  These are
the steps in which the NASH algorithm accepted a new baseline from
which the learning then proceeded.

\begin{figure}
\centering
\includegraphics[width=0.49\textwidth]{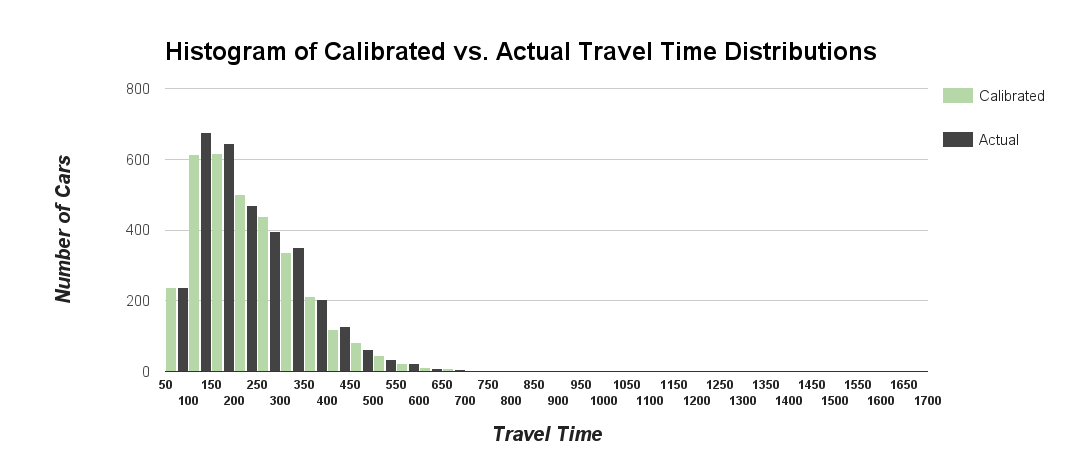}
\hfill
\includegraphics[width=0.49\textwidth]{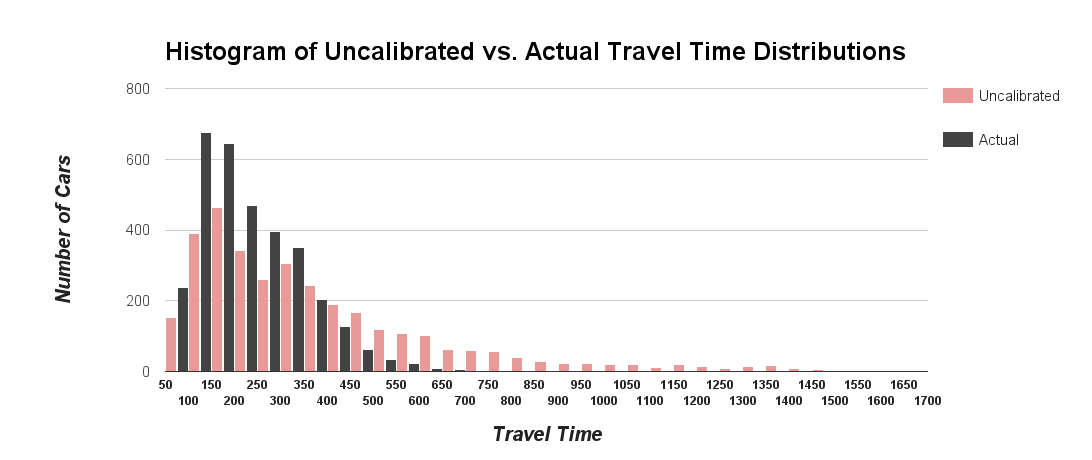}
\caption{ Distribution of travel times for 3200 cars.  (Top)
  Lights calibrated with NASH.  (Bottom)
  uncalibrated lights.  These are lights with random phases and
  offsets --- presented for comparison. }
\label{fig:dists}
\end{figure}

To compare the distributions of travel times to the actual travel
times, see Figure~\ref{fig:dists} (Top).  The actual travel time
distribution is shown with dark bars, the travel time distribution
obtained with the calibrated light settings is shown with lighter
(green) bars. As can be seen, the distribution of travel time appear
similar.  For comparison, if we plot the travel times for random light
settings (uncalibrated), the distributions appear quite different
Figure~\ref{fig:dists} (Bottom).

\begin{figure}
\centering
\includegraphics[width=0.4\textwidth]{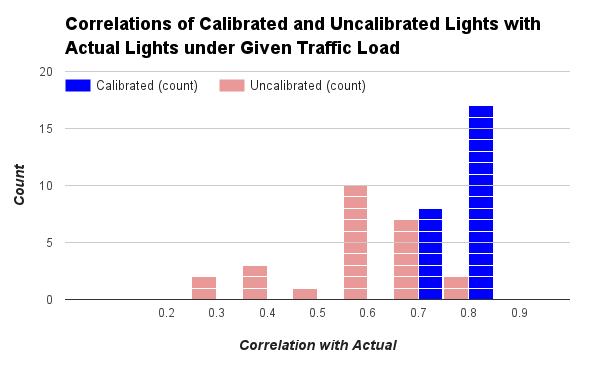}
\hfill
\caption{Correlation of calibrated and uncalibrated lights with
  actual travel times. Average calibrated correlation = 0.8, average
  uncalibrated correlation = 0.5. }
\label{fig:multipleCorrelation}
\end{figure}

We can also measure the correlation of journey times under the
hypothesized light system and the target light system.  Here, we
correlate each car's travel times under both scenarios.  Results are
shown in Figure~\ref{fig:multipleCorrelation}.  Instead of just
correlating a single trial, Figure~\ref{fig:multipleCorrelation} also
shows the results of 24 additional tests.  We reran the entire
NASH-matching algorithm from scratch 25 times.  Because NASH is
stochastic and is initialized with random seeds, we fully expect
different schedules to be found in each run; the hope is that at the
end of each run, the aggregate behavior approximates the target light
system --- even if the light settings themselves are not the same.
The correlations to the actual travel times of all 25 calibrated
systems (found through NASH) remains high for all trials.

For comparison, also shown in Figure~\ref{fig:multipleCorrelation}
are the correlation of 25 uncalibrated (random) light systems to
actual travel times.  It might seem surprising that for random systems
there is \emph{any} correlation; however, since we are measuring travel
times, even with random light settings, as cars travel through the
entire system, longer paths are likely to have longer travel times,
regardless of the light settings.  Nonetheless, as can be seen, by
matching the lights through NASH, the correlation of travel times
increases dramatically.

\begin{figure}
\centering
\includegraphics[width=0.4\textwidth]{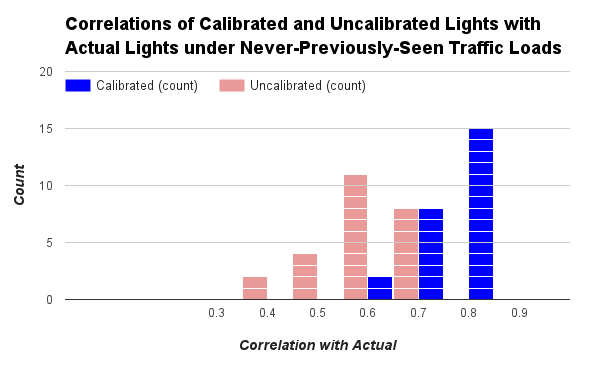}
\hfill
\caption{Correlations of calibrated and uncalibrated Lights to actual timings when all of the routes are replaced with never previously seen routes.}
\label{fig:modifiedLoad}
\end{figure}

Next we posit the question: how robust are these matches?  What
happens when we severely alter the underlying traffic profile?  If we
change all the routes and their distributions, how do the travel times
of the new cars compare in the NASH-calibrated light systems to the
original \emph{Target-Light-Setting}?  If the NASH calibrated light
systems were truly close to the \emph{Target-Light-Setting}, we would
expect a high correlation to remain under any traffic load, not just
the load on which it was trained.\footnote{This additional test is
  only possible for the data in Section~\ref{estimation}, since we are
  using purely synthesized traffic, and we therefore have the actual
  traffic light settings that we are trying to discover with NASH.  In
  real usage (e.g. the next two sections) this test would not be
  possible without having ground-truth light settings.}  To test this,
all of the routes are replaced with randomly created new routes.  We
measure the correlation of timings of the cars under the
target-light-schedule and the NASH-derived-light schedules, and the
uncalibrated light schedules (random).  The results are shown in
Figure~\ref{fig:modifiedLoad}.

The results are positive: under both seen
(Figure~\ref{fig:multipleCorrelation}) and unseen traffic loads
(Figure~\ref{fig:modifiedLoad}), after NASH is used to estimate the
settings of the lights, we observe that the correlation of travel
times to the original \emph{Target-Light-Setting} remains high.

\begin{figure}
\centering
\includegraphics[width=0.45\textwidth]{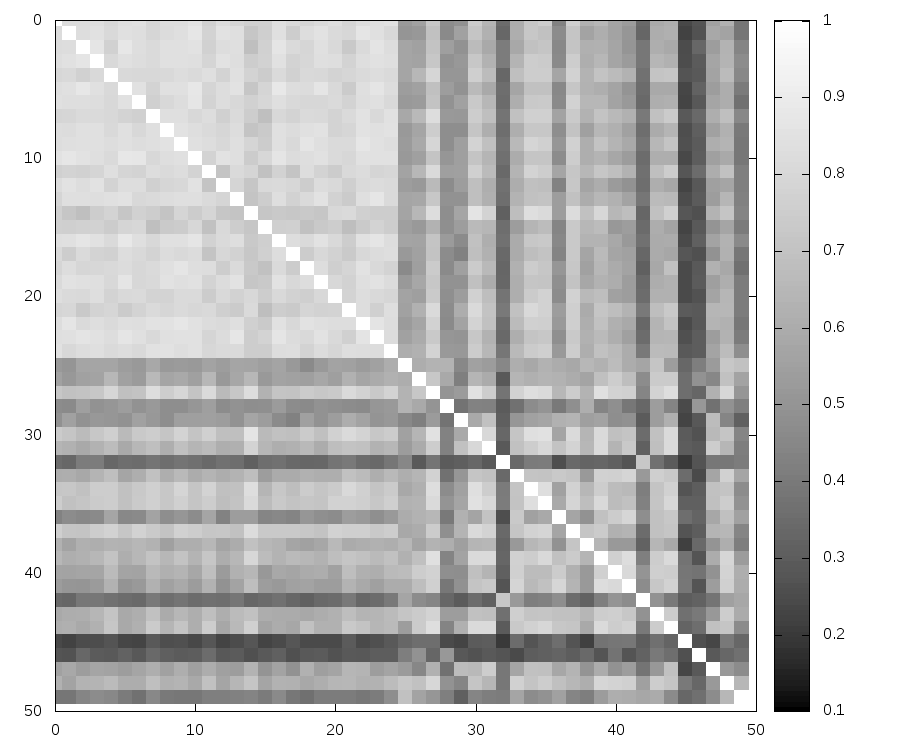}
\hfill
\caption{Correlations of $25\times$calibrated and
$25\times$uncalibrated lights to each other and across sets.  The
lighter quadrant (upper-left corner) is the 25 calibrated lights
compared to each other.  They exhibit a large degree of correlation.
The darker quadrant (lower-right corner) is the uncalibrated lights,
exhibiting far less correlation.  The white diagonal line is each
setting's correlation with itself (1.0).}
\label{fig:intensity}
\end{figure}

As a final test, we visualize how correlated the light settings are
with \emph{each other}.  If they all capture the same conditions well,
they should exhibit high correlation to each other.  We also compare
how correlated the 25 uncalibrated light settings are to both the
calibrated lights and uncalibrated lights.  The results are shown in
Figure~\ref{fig:intensity}. The upper-left quadrant (brighter) shows
the 25x25 correlations with the calibrated lights.  As can be seen,
they exhibit high correlations with each other.  When they are
compared to the non-calibrated lights, the correlations drop
precipitously. The bottom-right corner shows the correlations between
the uncalibrated lights; far less correlation is present.

In summary, the results are what we hoped: despite not being optimized
to be exactly the same light-programs (since the true light program
may not be known), the \emph{behavior} of the calibrated lights are
highly correlated with each other and also highly correlated to the
actual observed timings.  This is precisely what is required to obtain
a realistic baseline performance.

Next, we return our attention to optimizing the traffic lights to
improve capacity and throughput.  To measure our performance, we use the methods
described in this section to establish credible baseline performances
in both cities examined.

\section {Results: Mountain View, California}
\label{mtv}

In this section, we examine the performance of static lights,
long-term planning based lights, and auction based lights, on real
traffic from Mountain View, California.

\subsection{Calibrating to Real Lights}

The results from Section~\ref{estimation} revealed that matching the
behavior of lights was possible, at least with simulated data. In this
section, we apply the same approach to the seven traffic lights in
Mountain View, California.  This is a significantly larger and more
complex simulation than the synthetic data use earlier.  In the
previous section, the synthetic simulations used 3200 travel-tracks.
For these experiments, we use approximately 67,000 tracks.
Additionally, unlike in the synthetic experiments, the distribution of
paths is far from uniform.  Backups happen non-uniformly --- only on
certain streets and in certain directions.  Small perturbations in a
single critical traffic light's schedule can lead to drastic changes
in throughput while large perturbations to the schedule of a less busy
traffic light may generate little observable impact.

\begin{figure}
\centering
\includegraphics[width=0.6\textwidth]{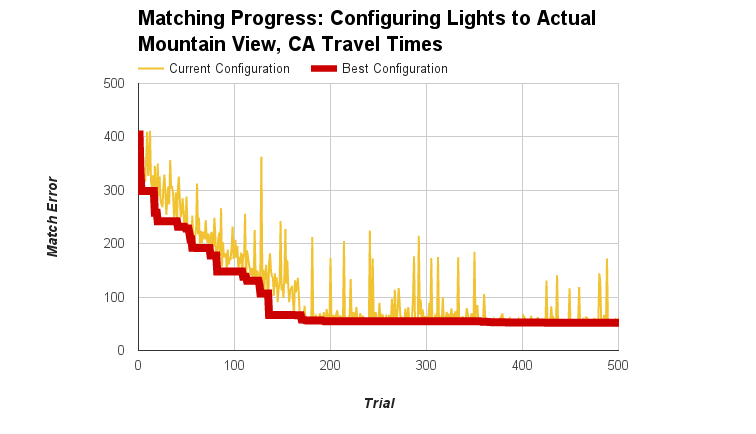}
\caption{ Progress on matching the real traffic using NASH.  Y-axis
  the average difference in journey times for cars under the
  hypothesized light setting and the real-data (in seconds).  X-axis:
  trial number.  First 500 trials shown - the remaining 1500 trials
  (not shown) showed continued, but smaller magnitude, improvements.}
\label{fig:mtvprogress}
\end{figure}

We begin the calibration procedure similarly to the synthetic case ---
we use NASH to match the timings.  The progress is shown in
Figure~\ref{fig:mtvprogress}.  Note that the mean error in times drops
quickly.  In the beginning of the learning process, the error was
over over 400 seconds (in this case, the default SUMO traffic light
settings were used for initialization).  By completion, NASH lowered
the error between actual and matched times to approximately 51
seconds.

\begin{figure*}
\centering
\includegraphics[width=0.9\textwidth]{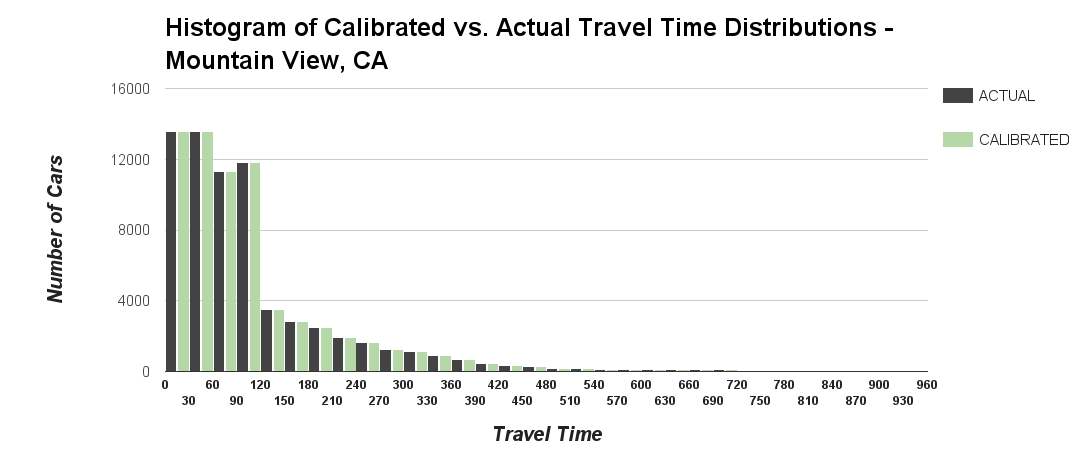}
\hfill
\caption{ Distributions of Real (Dark) and NASH-calibrated light travel times (Light/Green).}
\label{fig:mtvdist}
\end{figure*}

Next, similar to the analysis conducted with synthetic data, we
examine the distribution of travel times for the 67,000 tracks.
Figure~\ref{fig:mtvdist} shows the distributions of the tracks for the real
data and the times obtained through a simulation in SUMO using the
NASH-calibrated lights.

As can be seen the distributions are similar, but as expected, not as
close a match as with the simulated data.  The correlation between the
predicted and real times is 0.5.  For reference, when a random light
setting is used (the default SUMO settings), the correlation with the
real data is only 0.18.  Recall that with the synthetic data, the
correlation of NASH-Calibrated lights to the actual timings was 0.8
and the random light settings to the actual timings was 0.5.

There are several distinguishing attributes of this data that explain
why it was possible to correlate the synthetic data better than the
real data.  First, in the synthetic data, drivers did not have
unnecessary delays in start/stop times due to exogenous factors such
as distraction, change of plans, etc.  Second, in the synthetic data,
drivers behaved uniformly at yellow and red lights; this was not the
case for real drivers.  Even if the exact correct light settings were
found, these factors would lead to lower correlations since they would
not be modeled.  Third, the real data is significantly more noisy; a
problem we did not have with synthetic data.  Recall how the data was
acquired and aggregated over the period of many months (done to
account for the fact that not all drivers are opted-in Android phone
users).  This aggregation process leads to noise in the real data
which we are trying to mimic.  Fortunately, as usage of cell-phone
GPS/maps increases and more travel-track data becomes available, these
estimates will improve.  Despite the above difficulties, we were able
to significantly reduce the average discrepancy between actual and
predicted times from approximately 7 minutes to 51 seconds.

As a final test to ensure that NASH actually learned something about
the lights and did not overfit the exact traffic on which it was
trained, we repeated the experiment with a more difficult setup.  For
\emph{NASH}, we only looked at the data from the first 1.5 months of
collection.  Then, for \textbf{testing} the timings we tested on a
non-overlapping, subsequent, 1.5 months of data.  This also represents
another common use and test scenario where a period of time is devoted
to training and then the learned model is used in the future.  What we
found indicated that NASH captured the behavior of the lights and did
not overfit the data.  The results did not change from using the
entire 3 months of data. Both the mean error and overall correlation
remained the same as when the entire 3 month period was used for
modeling the lights.

\subsection{Testing the New Traffic Light Controllers}

Thus far, we have used NASH to match the timings of the lights in
simulation to those actually deployed on the roadways.  Now, we use
NASH in a manner that is more familiar --- to optimize the internal
parameters of the light controllers to reduce the mean travel times.
NASH is used to optimize the parameters of the three types of
controllers.  For fixed-schedule controllers, the phase lengths and
offsets are optimized, for long-term planning controllers the penalty
weights are optimized, and for micro-auction based controllers the
detector weights and their durations (minimum, priority and release)
are optimized.  The objective function (step 3 in the NASH algorithm)
is set to minimizing the mean travel time of the cars.  Each learning
run is given a total of 2000 simulations -- that is, 2000
perturbations of the traffic controller's parameters were tested in
simulation.

The traffic-data used for the learning process were based on the
data collected over months in Mountain View, California.  Rather than
using this exact data for the learning, multiple new,
\emph{reality-based}, data sets were created based on this data.  Each new
data set was created by perturbing the original data in the number of
cars released on each route and the car-release schedules.  Small
changes in the order of car release or in the number of cars can have
very large effects on the overall traffic and timings of the cars in
the simulation.  Therefore, instead of optimizing the signal timings
on a single data set, simulations were run for all the data sets
(10 dataset perturbations were created and used).  The
lights were optimized to work well on \emph{all} the data sets by
incorporating them into the objective function. During the
learning  process, to determine whether a candidate light setting
was better than its predecessor (Steps 4 \& 5 in the NASH algorithm),
not only did the average travel times across data sets have to be
lower than its predecessor, but of the $N$ data sets that were used,
at least $N/2$ had to have a lower mean travel times.  This ensured
that if a candidate light setting worked exceedingly well on a single
data set at the expense of working well on the rest, it was discarded,
despite potentially having a lower overall average across all data
sets.

The benefits of using the multiple \emph{reality-based} data sets were
two-fold.  First, multiple data sets helps account for naturally
occurring variability in traffic patterns on different days.  Since
small changes cause large distortions in traffic flow, this is vital.
Second, by not using the actual data for training the signals, this
left the pristine, real, data for the final round of tests.  The clean
data was used only as a final test to measure effects of improved
controllers on capacity and mean time of travel.

The results, comparing mean travel times, and expected capacity, are
shown in Tables~\ref{speed comparisons} and~\ref{capacity
comparisons}, respectively.  We compare different approaches to
traffic light control during different commuter-traffic profiles.  We
perform the comparison by considering as our baseline the travel times
under the NASH-calibrated light controls.

\begin{table*}
    \small
    \centering
    \caption{Mean-travel-time (MTT) changes under matched demand - Mountain View, California}
    \begin{tabular}{p{4cm}llllll}
    \toprule
    & \multicolumn{2}{p{4cm}}{Peak rush (8:00--9:30 am)} & \multicolumn{2}{p{4cm}}{Average rush (9:00--10:30 am)} & \multicolumn{2}{p{4cm}}{Low rush
    (9:30--11:00 am)} \\
    & \multicolumn{2}{p{4cm}}{21861 observed cars} & \multicolumn{2}{p{4cm}}{20139 observed cars} & \multicolumn{2}{p{4cm}}{18154 observed cars} \\
    & \multicolumn{2}{p{4cm}}{668.68 sec observed MTT} & \multicolumn{2}{p{4cm}}{173.65 sec observed MTT} & \multicolumn{2}{p{4cm}}{110.37 sec observed
MTT} \\
    \midrule
    Optimized static lights & 44\% faster & (376.56 sec MTT) & 30\% faster & (121.37 sec MTT) & 6\% faster & (103.58 sec MTT) \\
    Planning-based lights & 21\% faster & (525.82 sec MTT) & 8\% slower & (187.18 sec MTT) & 6\% slower & (116.54 sec MTT) \\
    Learned Auction lights & 79\% faster & (140.87 sec MTT) & 38\% faster & (108.01 sec MTT) & 10\% faster & (99.57 sec MTT) \\
    \bottomrule
    \end{tabular}
    \label{speed comparisons}

\end{table*}

\begin{table*}
    \small
    \centering
    \caption{Capacity changes under matched MTTs - Mountain View, California}
    \begin{tabular}{p{4cm}llllll}
    \toprule
    & \multicolumn{2}{p{4cm}}{Peak rush (8:00--9:30 am)} & \multicolumn{2}{p{4cm}}{Average rush (9:00--10:30 am)} & \multicolumn{2}{p{4cm}}{Low rush
    (9:30--11:00 am)} \\
    & \multicolumn{2}{p{4cm}}{21861 observed cars} & \multicolumn{2}{p{4cm}}{20139 observed cars} & \multicolumn{2}{p{4cm}}{18154 observed cars} \\
    & \multicolumn{2}{p{4cm}}{668.68 sec observed MTT} & \multicolumn{2}{p{4cm}}{173.65 sec observed MTT} & \multicolumn{2}{p{4cm}}{110.37 sec observed
MTT} \\
    \midrule
    Optimized static lights & +16\% & (25306 cars) & +9\% & (22037 cars) & +8\% & (19661 cars) \\
    Planning-based lights & +3\% & (22473 cars) & -4\% & (19253 cars) & -2\% & (17736 cars) \\
    Learned Auction lights & +47\% & (32082 cars) & +46\% & (29499 cars) & +11\% & (20204 cars) \\
    \bottomrule
    \end{tabular}
    \label{capacity comparisons}

\end{table*}

When comparing an alternative approach (e.g., optimized static-phase
controls) to the baseline, we run the simulation using the alternative
control approach, always using the same \emph{distribution} of routes
as was used for the baseline.  For Table~\ref{speed comparisons}, we
keep the same number of cars as well and simply compare the mean
travel-time (MTT) changes.  For Table~\ref{capacity comparisons}, we
scale the \emph{number} of cars up or down (with the same distribution
of routes), until we match its average travel time to the baseline
travel time.

The first lines of Tables~\ref{speed comparisons} and~\ref{capacity
  comparisons} report the results of our comparison of optimized
static control to the NASH-calibrated control.  The optimized static
control uses the phase durations and relative offsets that were found
to be best for an average-morning commute period (weekday 7am--11am),
using NASH.  There is improved throughput throughout the commute
cycle, with the largest improvements seen when it is most needed,
during the 8:00-9:30am peak period.  For example, we could increase
the peak period load by as much as 16\% without increasing the travel
times, using this learned static controller instead of the
NASH-calibrated controller.  This improvement demonstrates the power
of the learning procedure.  It is interesting to see this level of
improvement using static light timing, even after the light
synchronization work was done on Shoreline Blvd. in
2012--13~\cite{MTC2012}.  It may be that the increase in traffic in
those two years is enough to change the best choices for the traffic
light durations.

A potential concern with the witnessed improvements is whether the
improvements come at the expense of infrequently traveled routes
experiencing severe delays that they previously did not?  These
trade-offs were examined in detail.  The vast majority of cars reduced
their travel times.  Of the ones that did not, most increased their
travel times by under 4 seconds.  Only a few cars experienced longer
delays under the optimized controllers; however, even with the extra
delays, their resulting times were in the lowest 10\% of the times
experienced without learning.

For planning-based lights (Table~\ref{capacity comparisons}), we did
not see a consistent change in capacity across the commute periods:
the implied changes in capacity, compared to the NASH-calibrated
control, were all small and likely not to be significant.  This is
most probably due to the situation that was studied.  The majority of
the traffic that was involved in the simulation were cars exiting from
a freeway onto a congested arterial road.  Most of the traffic was not
in ``clusters'', as observed in the Pittsburgh study~\cite{Smith2013}.
Another potential contributing factor is that there were only three
parameters (the penalty weights) that were optimized.  We made this
choice because of the obvious physical correlates for all of the other
parameters (\emph{e.g.}  observed speed profiles for sensor-to-light
delays and turning ratios for sensor-to-phase weights).

Finally, our auction-based traffic lights provided the largest gains
in capacity over the matched-to-current controls.  The biggest gains
were during the peak rush period, when it is most needed: \textbf{we
  were able to allow 47\% more traffic into the road network} (using
the same distribution of routes) during these times, \textbf{without
  increasing travel times}, when we used auction-based traffic
controls compared to the matched-to-current lights, a significant
accomplishment.  In contrast, if we increased the traffic load during
peak hours by 47\%, the currently deployed lights (as modeled by the
procedures described in Section~\ref{estimation}), we expect to see
the travel times increase by over 200\% (to ~2203.6 sec).  Similarly,
impressive reduction in travel times were achieved across the board,
especially during peak rush hour.

\subsection {Detailed Discussion}

Because of the stochastic nature of NASH, as well as the complex
interactions possible between a system of traffic lights and real
traffic, it is difficult to \emph{a priori} predict what behaviors will emerge after
learning.  From the previous tables, it was clear that
throughput and mean travel time improvements were found.  In this
section, we attempt to give some intuition into (the sometimes
non-obvious) ways in which they were achieved.

Somewhat surprisingly, the planning-based approach was the least
successful of the systems that we tested.  Due to the strong physical
interpretation of each of the parameters (physical distances, etc), it
was the one that least lent itself to optimization.  A future research
direction is to test this approach using fully optimized parameters
that may not correspond to physical interpretations, but may reveal
improved performance within the context of the full system.

Examining the auction-based lights in more detail, the decisions
learned are both interesting and revealing.  In four of the seven
lights, the learning procedure removed all sensors from all the
phases, resulting in the lights acting as static lights: lights~1, 2,
5, and~7 (see Figure~\ref{fig:mtv-sumo} for numbering).  The phase
durations seen on all of these lights except light~5 were nearly
identical to the durations found when we optimized static-control
lights: the other three were within 2 sec (out of 86 sec total cycle
length) of the optimized static-control light durations.  At least
this small amount of variation (if not more) is expected from a
stochastic search process, such as NASH.

The learning procedure  also found several different ways to, in
effect, remove dedicated-left phases, especially at the most congested
intersections (for example, lights~3 and~6 in
Figure~\ref{fig:mtv-sumo}).\footnote{The largest amount of congestion
  is at light~4.  However, due to the complexity of that intersection,
  there was no dedicated-left phase to remove.} This behavior seems
especially interesting, given the reduced capacity that is generally
seen at intersections with many dedicated lefts~\cite{Vanderbilt2009}.

One approach that the system automatically discovered to remove
dedicated lefts from the less-used roads was to match the bids of the
dedicated left to the direct-through phase that was just before it, in
the round-robin ordering.  Specifically, on light~6
(Figure~\ref{fig:mtv-sumo}), phases~1 and~2 share a matched set of
sensor weights and phases~3 and~4 share another matched set of sensor
weights.  This means that the direct-through/dedicated-left pair will
always tie on any auctions.  Since the direct-through phases are
always before the dedicated left phases in the round-robin cycle when
starting from an opposing direction (e.g., phase~1 will be before
phase~2, when starting from either phase~3 or phase~4), this has the
interesting property that cycling between directions (cycling from
north-south to east-west or vice-versa) will \emph{always} go to the
through phase.  

In visual examination of this intersection during simulation, it often
happened that phases~1 and~2 started giving high bids for the light,
while phase~3 (east-west direct-through) was active but still greedy.
As soon as phase~3 is slightly longer than its priority duration, the
light changes to using phase~1 (north-south direct-through), without
going through the phase~4 (the dedicated left).  Similar switching is
only occasionally seen for the opposite pairs.  Since Shoreline has
much heavier traffic than Terra Bella (the east-west road), this
happened much less often.  Both sets of dedicated lefts happen less
often but are still possible.

The learned controllers also used the time differences between
release durations to favor traffic on the phases that were more prone
to long lines.  For example, at light~4 (Figure~\ref{fig:mtv-sumo}),
the priority duration given to the 101-Freeway exit ramp was
$10\times$ longer than the priority duration given to La Avenida (the
small road that enters the intersection from the east).  With this
$10\times$ weighting, we see the traffic delays suffered by the La
Avenida traffic approximately equal (on a per-car basis) to the delays
seen by the 101-exit-ramp traffic.

Finally, the learned lights used the number of incoming lanes allowed
to pass through a phase as another way to bias the auction towards the
dominant direction of traffic.  For example, at light~3, the
\emph{combined weight} given to the sensors of Shoreline traffic is
five times that given to the Pear-Ave traffic (the east-west road at
that intersection).  This was done by giving all of the sensors the
maximum-allowed positive or negative weights and simply relying on the
number of sensors (and hence the number of lanes) to provide priority
to larger streets.

\section {Results: Chicago, Illinois}
\label{chicago}

\begin{figure}
\centering
\includegraphics[width=0.40\textwidth]{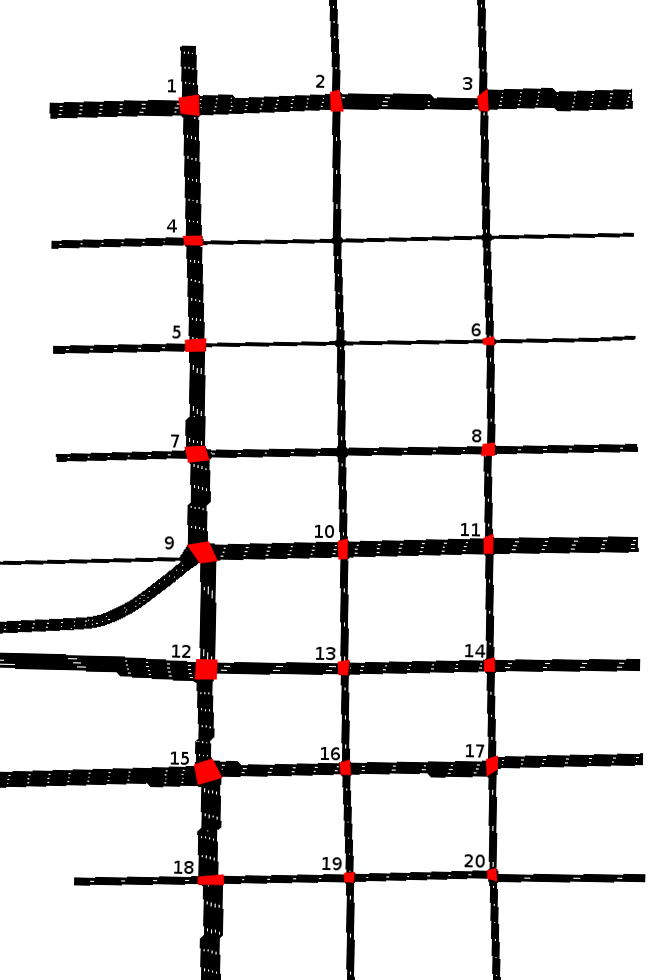}

\caption{ Chicago roadways as imported in to SUMO.  The twenty lights
  to be optimized are marked.}
\label{fig:chicagosumo}
\end{figure}

In Chicago, we considered twenty lights in the River North area shown
in Figure~\ref{figmtv} (Bottom).  The area, as imported into SUMO, is
shown in Figure~\ref{fig:chicagosumo}.

The steps in the optimization of the twenty traffic lights in Chicago
proceeded in the same manner as the optimization of the lights in the
Mountain View area.  To avoid redundancy, we will provide only a
summary of the procedure and results and refer the reader to
Section~\ref{mtv} for details.

\subsection{Calibrating to Real Lights}

In order to be able to establish a realistic baseline, as we did with
Mountain View, we once again used NASH to calibrate the lights in our
simulations (see Section~\ref{data}).  Similarly to the previous
experiment, anonymized, fully privacy preserving, travel-tracks of
Android users were collected over several months.  For our tests,
approximately 18,000 cars were used in the simulation.  By using NASH
to adjust the lights schedules, we were able to reduce the average
discrepancy between the actual and the predicted travel times to
approximately 43 seconds.  For comparison, in the analogous procedure
using the Mountain View, California data, the error was reduced to 51
seconds (see Figure~\ref{fig:mtvprogress}).

\subsection{Testing the New Traffic Light Controllers}

Once the settings for the deployed lights are estimated, the next step
is to attempt to change the light settings to \emph{improve} the
traffic flow.  In the first experiment, NASH is employed to modify the
phase durations and offsets of the 20 lights (\emph {e.g.} optimize
static lights).  In the second experiment, NASH is used to optimize
the parameters of the micro-auction based controllers.  Unlike the
previous experiments with Mountain View, CA.  we do not revisit the
planning based approach for this scenario.

\begin{figure}[t]
\centering
\includegraphics[width=0.49\textwidth]{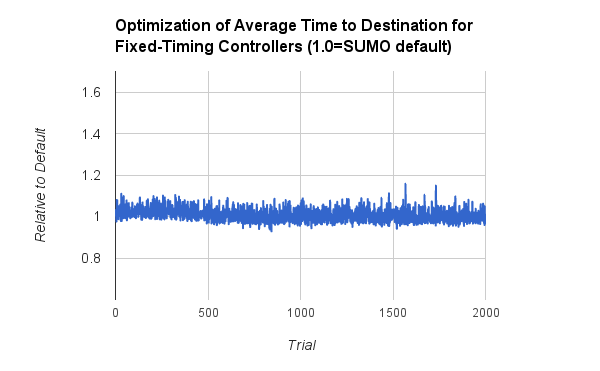}
\includegraphics[width=0.49\textwidth]{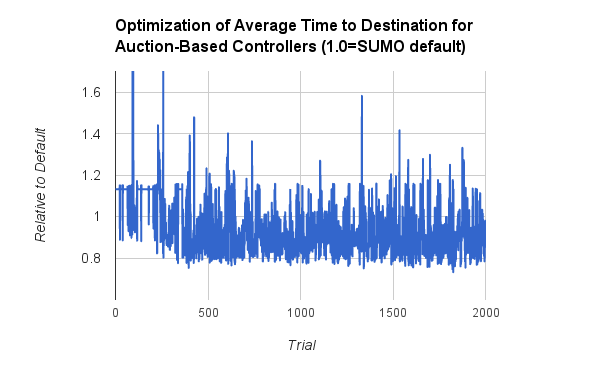}

\caption{Progress of average time of travel as learning of controller
  parameters occurs for the 20 lights in Chicago.  \emph{Lower is
    better}.  1.0 on the Y-Axis is the base performance. Variations
  from 1.0 are due to perturbations in the controller's settings.
  TOP: Optimization of fixed light schedules.  BOTTOM: Optimization of
  micro-auction based traffic lights.  Each run is allowed 2000
  evaluations (controller settings) to try. Note that many controller
  settings result in performance worse than the initial setting; these
  are discarded.  Also note that the learning of micro-auction
  controllers allows the average time of travel to drop substantially
  more than optimizing the fixed-time controllers.}

\label{fig:chicagoopt}
\end{figure}

Learning, as it progresses, is shown in Figure~\ref{fig:chicagoopt}.
From the two graphs, we see several important trends.  First,
optimizing the fixed-light controllers (Top) does not dramatically
improve the performance of the fixed-light controllers in terms of the
average time to destination.  Compared to the default settings, the
best setting found revealed an improvement of approximately 7\%.

In contrast, there are much larger variations in the optimization of
the auction based controllers.  In Figure~\ref{fig:chicagoopt}
(Bottom), we see that in the 2000 trials, many times the settings are
significantly worse than the original.  As explained earlier, these
settings are tried and then discarded.  Most importantly, however, in
the cases in which better controller settings are found, a substantial
improvement over the fixed-light controllers is obtained.  With the
auction-based light controllers, the learning process discovered
settings that reduced the mean travel time by 25\%.

\begin{table*}
  \centering
  \small
    \caption{Chicago: Mean-travel-time (MTT) changes under matched demand}
    \begin{tabular}{p{4cm}llllll}
    \toprule
    & \multicolumn{2}{p{4cm}}{Peak rush (8:00--9:30 am)} & \multicolumn{2}{p{4cm}}{Average rush (9:00--10:30 am)} & \multicolumn{2}{p{4cm}}{Low rush
    (9:30--11:00 am)} \\
    & \multicolumn{2}{p{4cm}}{7516 observed cars} & \multicolumn{2}{p{4cm}}{6834 observed cars} & \multicolumn{2}{p{4cm}}{6390 observed cars} \\
    & \multicolumn{2}{p{4cm}}{87.18 sec observed MTT} & \multicolumn{2}{p{4cm}}{86.42 sec observed MTT} & \multicolumn{2}{p{4cm}}{85.71 sec observed MTT} \\
    \midrule
    Optimized static lights & 2\% faster & (85.63 sec MTT) & 3\% faster & (84.02 sec MTT) & 3\% faster & (83.17 sec MTT) \\
    Learned Auction lights & 24\% faster & (66.40 sec MTT) & 23\% faster & (66.57 sec MTT) & 25\% faster & (64.41 sec MTT) \\
    \bottomrule
    \end{tabular}
    \label{speed comparisons chicago}

\end{table*}

\begin{table*}
  \centering
  \small  
    \caption{Chicago: Capacity changes under matched MTTs}
    \begin{tabular}{p{4cm}llllll}
    \toprule
    & \multicolumn{2}{p{4cm}}{Peak rush (8:00--9:30 am)} & \multicolumn{2}{p{4cm}}{Average rush (9:00--10:30 am)} & \multicolumn{2}{p{4cm}}{Low rush
    (9:30--11:00 am)} \\
    & \multicolumn{2}{p{4cm}}{7516 observed cars} & \multicolumn{2}{p{4cm}}{6834 observed cars} & \multicolumn{2}{p{4cm}}{6390 observed cars} \\
    & \multicolumn{2}{p{4cm}}{87.18 sec observed MTT} & \multicolumn{2}{p{4cm}}{86.42 sec observed MTT} & \multicolumn{2}{p{4cm}}{85.71 sec observed MTT} \\
    \midrule
    Optimized static lights & +9\% & (8195 cars) & +26\% & (8601 cars) & +31\% & (8362 cars) \\
    Learned Auction lights & +26\% & (9505 cars) & +95\% & (13360 cars) & +150\% & (15965 cars) \\
    \bottomrule
    \end{tabular}
    \label{capacity comparisons chicago}

\end{table*}

As with the study conducted with the Mountain View, California, the
settings for the light controllers are found by training with multiple
\emph{reality-based} data sets. This helps to account for variability
in traffic flows as well as leaves the actual data untouched for
testing.  For the final results, reported below, we use the actual
data.

Tables~\ref{speed comparisons chicago} and~\ref{capacity comparisons
  chicago} compare the mean-travel times and the capacity changes to
NASH-calibrated lights.  As witnessed in Mountain View, both the
optimized static and the auction-based traffic lights provide
improvements in the mean-travel times and in the capacity of the
streets: up to 25\% faster mean-travel time for the auction-based
traffic lights and, during the main commute hours, up to 150\% more
capacity during low rush hours.

The traffic problems addressed in this region of Chicago are different
than those in Mountain View, California.  In Mountain View, most of
the traffic delay was associated with long, persistent, queues at
intersection 4 in Figure~\ref{fig:mtv-sumo} (Shoreline and the
northbound exit from 101 and 85) and with the addition of traffic onto
Shoreline from that exit.  In the Chicago study, there is a busy
freeway exit at intersection 12 in Figure~\ref{fig:chicagosumo}, but
the traffic clears from the ramp on most light cycles.  Instead, the
delays in traversing this section are largely due to repeated stops at
the traffic lights along W. Ohio (intersections 12, 13, and 14 in
Figure~\ref{fig:chicagosumo}) and along W. Ontario (intersections 9,
10, and 11).

The optimized static lights are able to increase capacity and decrease
travel times by changing the phase duration allocation to better match
the dominant directions of congestion at the different intersections,
most significantly increasing phase durations for east-west traffic as
it crosses N. Wells and N. Franklin and increasing the phase duration
for north-south traffic at N. Orleans and W. Ontario (intersection 9)
by 14\%.  The phase offset between the lights was unchanged from the
matched lights to the optimized static traffic lights.  This helps
explain why the travel times had a smaller improvement than the
capacity measurements: many cars still slowed or stopped for the same
sets of lights, but the length of the queues were smaller (even with
increased traffic load), due to the optimized match to the
historically-observed congestion.

The explanation for the improved capacity for the actuated lights is
similar but more indirect.  We cannot simply examine the phase
durations, since the duration is demand driven and changes throughout
the simulation.  Furthermore, the average change in phase duration
from matched to actuated lights (that is, the difference in the
percent time that a given phase was green over the full simulation
run) was not consistent across the different time windows.  The most
noticeable and consistent change between the static and auction-based
light simulations was that the size of the queues for the major
intersections were shorter in all directions using the actuated
traffic lights.  The first car to the intersection would typically
still need to slow or stop but, the following cars in the same wave
would typically make it through the same cycle.  Examining the way
that the sensors were used by the actuated lights, there seems to be a
bias towards using all of the sensor readings to establish the bid of
the phase for the dominant direction of travel (intersections 1, 2, 3,
9, 11, 12, 14, 15), with the weighting on the sensors leading the bid
to pay more attention to traffic in the dominant direction but to
yield the light if that traffic is less than seen on the non-dominant
direction sensors.

Interestingly, as with many of the optimized Mountain View lights,
most of the protected-left phases were effectively removed from the
light sequence.  In most cases, this was done by having the
through-traffic phases with sensor weights that are ``matched
opposites'' (that is, all the sensor weights for phase 3 are the
negative of the weights for phase 1, in Figure~\ref{intersection}).
Here, the protected-left can be effectively removed by omitting
sensors from that phase (lights 1, 3, 4, 7, 9, 16, 19) or by matching
the sensor weights for the protected left to the previous
through-traffic phase (lights 2, 3, 7, 9, 15, 16, 19).\footnote{Lights
  3, 7, 9, 16, and 19 appear on both of these lists since their two
  protected-left phases used both of the strategies.}

\section {Conclusions and Future Work}
\label{concs}

There are two primary contributions of this work.  First, we have
introduced the micro-auction based traffic light controllers and a
simple learning/optimization procedure to make them effective. The
controllers do not necessitate communication between lights or between
cars and lights; only local induction loops sensors are used.  Rather
than coming up with ad hoc rules for traffic priorities and
scheduling, each sensor's information is placed within the framework
of a micro-auction system.  When a phase change is permitted, the
light controller collects bids from all the phases and conducts a
real-time micro-auction.

We tested the controllers on real traffic loads observed over months
of traffic collection.  In Mountain View, California, the load was
dominated by the confluence of a freeway exit ramp with a congested
arterial road. Next, we studied real traffic in Chicago, Illinois, to
determine the controller's performance in within-city grids.  For both
cities, we employed the exact same method to estimate the performance
of the deployed traffic lights with no changes to the algorithms.  We
successfully demonstrated that with auction-based controllers, the
improvements in capacity and mean travel times in both cities were
extremely promising.  Perhaps equally as interesting as the
quantitative results are the common ways in which the learning
procedures worked with the auction-based controllers in finding
similar strategies for improving traffic in both cities: eliminating
protected lefts, weighting bids higher from dominant roads, and even
negatively weighting bids from less traveled roads.  These strategies
were automatically determined; none were pre-programmed.

The second contribution of this work is a method through which the
program schedules of currently installed traffic light can be
approximated.  This addresses an important practical issue in
measuring the benefits of new traffic light controllers.  For the
traffic researcher, two classes of crucial data have become
increasingly available.  First, detailed maps of the streets and the
precise locations of the traffic lights is publicly available through
a number of sources.  Second, through the increased usage of personal
cell-phone based GPS systems, an enormous amount of travel-tracks have
been amassed.  What is often lacking, however, is detailed knowledge
of the existing traffic light schedules and traffic light response
behaviors.  This paper has presented a simulation-based approach to
approximate \emph{the behavior} of installed lights.  This provides a
method to create a solid baseline from which to quantitatively measure
improvements.

There are several areas of future research that are natural next
steps. First, with respect to our procedures used to determine the
schedules of the currently deployed lights, in this study, we only
considered matching the timings of static-lights.  Nonetheless, it
should readily possible to adapt the same methods presented here to
learning the parameters for lights that incorporate induction loop
sensors.  We suspect that this will provide improvements even in the
scenarios considered here.

Second, in this study, the planning-based lights did not perform as
well as the auction-based system.  In the future, if planning-based
lights are reexamined, further parameterization of the sensor
placement and readings should be considered, despite the potential for
\emph{a priori} setting weights based on physical constraints.  This
may require new, more complex learning algorithms that explicitly
model the interactions between parameters; many have been explored in
genetic algorithm literature~\cite{baluja1998,pelikan2002,harik1999},
among other places.

Third, a large amount of research has been conducted towards lights
that communicate between each other and with cars.  If this form of
communication is available, it can be easily incorporated into auction
based models. In the simplest manner, the induction loop sensors can
be replaced with counters revealing the presence of cars through
communication channels.  However, more interesting interactions are
also possible.  If there is higher priority traffic (for example
emergency vehicles or lanes that need right-of-way for event traffic
management), in an auction based system their weights can be
appropriately accounted for in a phase's bidding process to reflect
the urgency of travel.

The focus of this paper was not in exposing the auction mechanics
externally, but rather using the auction as a guiding principle for
internal light controls.  In the future, if lights do communicate with
cars, it is conceptually easy to incorporate tolls and payment
priorities into our auction process.  If each car has the ability to
virtually ``bid'' on how much it is willing to pay to make it through
the lights, the auction-based mechanisms support this by altering the
weights of the bids that are received into the micro-auction.  Of
course, setting the weights and the algorithms to be fair will require
careful consideration; nonetheless, the mechanisms to support monetary
bidding are provided through these learned auctions-based controllers.

\section*{Acknowledgments}

The authors would like to gratefully acknowledge Kevin Wang for his
helpful comments on this paper and many aspects of this project over
the last year.

\section*{References}
\bibliography{itsc}

\end{document}